\documentclass[%
 aip,
 amsmath,amssymb,
 reprint,%
]{revtex4-1}

\usepackage{graphicx}
\usepackage{dcolumn}
\usepackage{bm}
\usepackage{booktabs}
\usepackage{mathtools}

\usepackage[utf8]{inputenc}
\usepackage[T1]{fontenc}
\usepackage{mathptmx}
\usepackage{etoolbox}

\makeatletter
\def\@email#1#2{%
 \endgroup
 \patchcmd{\titleblock@produce}
  {\frontmatter@RRAPformat}
  {\frontmatter@RRAPformat{\produce@RRAP{*#1\href{mailto:#2}{#2}}}\frontmatter@RRAPformat}
  {}{}
}%
\makeatother

\usepackage{xparse}
\usepackage{xcolor}
\usepackage{array}
\usepackage{colortbl}

\definecolor{rulegray}{gray}{0.75}
\definecolor{mygreen}{RGB}{0,150,0}

\usepackage[capitalize,noabbrev]{cleveref}

\usepackage[textsize=tiny]{todonotes}

\begin{document}

\title{Enhanced Diffusion Sampling: Efficient Rare Event Sampling
and Free Energy Calculation with Diffusion Models}

\author{Yu Xie}
 \altaffiliation{These authors contributed equally.}
\affiliation{Microsoft Research AI for Science}

\author{Ludwig Winkler}
 \altaffiliation{These authors contributed equally.}
\affiliation{Microsoft Research AI for Science}

\author{Lixin Sun}
 \altaffiliation{These authors contributed equally.}
\affiliation{Microsoft Research AI for Science}

\author{Sarah Lewis}
\affiliation{Microsoft Research AI for Science}

\author{Adam E. Foster}
\affiliation{Microsoft Research AI for Science}

\author{Jos\'e Jim\'enez Luna}
\affiliation{Microsoft Research AI for Science}

\author{Tim Hempel}
\affiliation{Microsoft Research AI for Science}

\author{Michael Gastegger}
\affiliation{Microsoft Research AI for Science}

\author{Yaoyi Chen}
\affiliation{Microsoft Research AI for Science}

\author{Iryna Zaporozhets}
\affiliation{Microsoft Research AI for Science}

\author{Cecilia Clementi}
\affiliation{Microsoft Research AI for Science}

\author{Christopher M. Bishop}
\affiliation{Microsoft Research AI for Science}

\author{Frank No\'e}
\affiliation{Microsoft Research AI for Science}

\date{\today}

\begin{abstract}
The rare-event sampling problem has long been the central limiting factor in molecular dynamics (MD), especially in biomolecular simulation. Recently, diffusion models such as BioEmu have emerged as powerful equilibrium samplers that generate independent samples from complex molecular distributions, eliminating the cost of sampling rare transition events. However, a sampling problem remains when computing observables that rely on states which are rare in equilibrium, for example folding free energies.
Here, we introduce enhanced diffusion sampling, enabling efficient exploration of rare‑event regions while preserving unbiased thermodynamic estimators. The key idea is to perform quantitatively accurate steering protocols to generate biased ensembles and subsequently recover equilibrium statistics via exact reweighting. We instantiate our framework in three algorithms: UmbrellaDiff (umbrella sampling with diffusion models), MetaDiff (a batchwise analogue for metadynamics), and $\Delta$G‑Diff (free‑energy differences via tilted ensembles).
Across toy systems, protein folding landscapes and folding free energies, our methods achieve fast, accurate, and scalable estimation of equilibrium properties within GPU-minutes to hours per system---closing the rare‑event sampling gap that remained after the advent of diffusion‑model equilibrium samplers.
\end{abstract}

\maketitle

\section{Introduction}
Molecular dynamics (MD) simulation is a widely used computational approach for generating molecular equilibrium ensembles $p(x)$ and predicting experimental observables $O=\mathbb{E}_{p(x)}[o(x)]$, but its effectiveness is limited by the sampling problem, which consists of two distinct components. 

\begin{enumerate}
    \item  \textit{Slow mixing problem}---MD produces time-correlated trajectories $x_t$. Long-lived states or phases lead to trapping the simulation trajectory for long times, resulting in slow exploration and slow convergence of expectation values.
    \item \textit{Rare state problem}---even with independent draws from $p(x)$, it can be prohibitive to sample states with small equilibrium probabilities. For example, the probability ratio of unfolded and folded protein states depends exponentially on the folding free energy: $p_u/p_f = \exp(\Delta G_{\mathrm{fold}}/k_BT)$. At $300$K, $\Delta G_{\mathrm{fold}}=-5$\,kcal/mol implies that $\sim$1 in $4.4\times10^3$ equilibrium samples is unfolded. For a moderately stable protein ($\Delta G_{\mathrm{fold}}=-10$\,kcal/mol) only $\sim$1 in $1.9\times10^7$ samples is unfolded. 
\end{enumerate}

These limitations have motivated enhanced sampling methods over the last 70 years \cite{Henin_Lelievre_Shirts_Valsson_Delemotte_2022}, which address rare states by sampling from a biased distribution and then reweighting to recover equilibrium statistics; however, when implemented on top of MD they can remain limited by slow mixing of the unbiased degrees of freedom. 
Recently, generative equilibrium samplers based on normalizing flows and diffusion models \cite{NoeEtAl_19_BoltzmannGenerators,bioemu2025} have emerged that generate approximately independent equilibrium configurations, removing the slow-mixing bottleneck, but rare-state estimation remains when observables depend on low-probability regions of $p(x)$. 
This paper develops a framework for enhanced sampling with diffusion-model samplers, addressing both bottlenecks within a single approach.

\begin{table}[t]
\caption{The MD sampling problem consists of a slow mixing problem due to rare interconversion between long-lived states, and a rare state problem as low-probability states are infrequently visited. Diffusion equilibrium samplers tackle the slow mixing problem, enhanced sampling methods tackle the rare state problem. In this paper we explore the combination of both: enhanced diffusion samplers.}
\label{tab:sampling_problem}
\renewcommand{\arraystretch}{1.2}
\setlength{\tabcolsep}{6pt}
\setlength{\arrayrulewidth}{0.8pt}
\arrayrulecolor{rulegray}
\begin{tabular}{l|l|l}
 & \textbf{Equilibrium sampling} & \textbf{Enhanced sampling} \\
\hline
\textbf{Molecular} & \textcolor{red}{Slow mixing problem} & \textcolor{red}{Slow mixing problem} \\
\textbf{dynamics} & \textcolor{red}{Rare state problem} & \textcolor{mygreen}{Rare states sampled} \\
\hline
\textbf{Diffusion} & \textcolor{mygreen}{Independent samples} & \textcolor{mygreen}{Independent samples} \\
\textbf{samplers} & \textcolor{red}{Rare state problem} & \textcolor{mygreen}{Rare states sampled} \\
 \hline
\end{tabular}
\end{table}

Traditional enhanced sampling methods include thermodynamic integration \cite{Kirkwood1935TI,Hummer2001JCPFastGrowthTI}, free energy perturbation (FEP) \cite{Zwanzig_JCP54_TI}, umbrella sampling \cite{Torrie_JCompPhys23_187,SouailleRoux_CPC01_WHAM,ZhuHummer_JCC12_ConvergenceWHAM}, parallel or simulated tempering \cite{Marinari1992,Hansmann1997,Sugita_CPL314_141}, metadynamics \cite{LaioParrinello_PNAS99_12562} and free energy reconstruction from nonequilibrium paths \cite{HummerSzabo2001PNAS,HummerSzabo2010PNAS}---see \cite{Henin_Lelievre_Shirts_Valsson_Delemotte_2022} for an extensive review. All these methods sample from a biased distribution, and then later remove this bias from the sampled statistics in order to recover equilibrium statistics \cite{Bennett_JCP76_BAR,FerrenbergSwendsen_PRL89_WHAM,ShirtsChodera_JCP08_MBAR}. 
They can accelerate sampling by orders of magnitude when suitable collective variables or thermodynamic controls are available. 
%
Representative successes include: (i) \textit{Free energy profiles of reactions and of ion permeations through channels} \cite{tokita2025strecker,stocker2025catalysis,Hub2008SelectivityAquaporins,Heer2017KcsASelectivityFilter}, where umbrella sampling can restrain sampling along the well-defined reaction coordinate, while the other degrees of freedom relax quickly; 
(ii) \textit{small-molecule solvation and protein--ligand binding free energies} \cite{cournia2017rbfe,moore2025mlsolvation,ChoderaEtAl_CurrOpin11_Alchemical}, where alchemical methods relying on free energy perturbation (FEP) connect nearby thermodynamic states with feasible local sampling;
(iii) \textit{small protein folding in implicit solvent} \cite{PiteraSwope_PNAS03_TrpCageREMD}, where replica exchange remains tractable; and 
(iv) \textit{mutation series in coarse-grained models} \cite{CharronEtAl_NatChem25_Navigating}, where reduced resolution makes otherwise prohibitive transitions amenable to free-energy calculations.

For high-dimensional biomolecular transitions---including protein folding, binding, and conformational changes in explicit solvent---enhanced sampling is often limited by two coupled issues. 
First, suitable low-dimensional bias coordinates are frequently unknown \emph{a priori} and may only become apparent after substantial sampling. This has motivated adaptive approaches that iteratively discover reaction coordinates and enhance sampling along them \cite{RibeiroTiwary_JCP18_RAVE,WangRibeiroTiwary2019_PFIB,ChenSidkyFerguson2019_SRV,MardtEtAl_VAMPnets,BonatiPicciniParrinello2021_DeepTICA_OPES,BonatiRizziParrinello2020_DeepLDA,KleimanShukla2023_MaxEntVAMPNet,PretoClementi2014_ExtendedDMdMD,ZhengRohrdanzClementi2013_DMdMD}. 
Second, these systems typically exhibit a \emph{spectrum} of slow relaxation processes rather than a single dominant timescale, as explored in the MSM literature \cite{PrinzEtAl_JCP11_MSM1,NoeEtAl_PNAS11_Fingerprints}; when slow modes are weakly separated, long simulations remain necessary to equilibrate degrees of freedom not directly controlled by the bias.

Consequently, successes on explicit-solvent biomolecular problems have often required specialized combinations of methods and/or massive compute. 
For \textit{all-atom protein folding} free-energy landscapes, temperature replica exchange becomes increasingly inefficient in explicit solvent because the number of replicas grows with system size, and practical studies have relied on hybrids such as REMD+metadynamics \cite{Juraszek2010,BajpaiAbreuNairTuckerman2025_STedAFED}, bias-exchange metadynamics \cite{Marinelli2009}, or multitemperature MD strategies \cite{Liu2012}, as well as special-purpose hardware or massively distributed simulations plus MSM analysis to obtain quantitative folding landscapes for proteins up to $\sim$100 residues \cite{LindorffLarsenEtAl_Science11_AntonFolding,PianaEtAl_PNAS13,BowmanVoelzPande_JACS11_FiveHelixBundle-TripletQuenching,Voelz2012}. 
For \textit{complex conformational changes}, well-characterized free-energy landscapes have been obtained using metadynamics and its variants \cite{Provasi2011PLOSCompBiol,Wang2016eLife17505}, string-based approaches combined with umbrella sampling or swarms of trajectories \cite{Meng2015JPhysChemB1443,Fleetwood2020Biochemistry}, and temperature-accelerated MD in collective variables \cite{AbramsVandenEijnden2010_TAMD_Proteins}. 
Large-scale unbiased simulation with MSMs has also enabled quantitative conformational landscapes in challenging systems \cite{KohlhoffEtAl_NatChem14_GPCR-MSM,Sultan2017BTKmsm,Bowman2012CrypticAllostery,ShuklaPande_NatCommun14_SrcKinase,Meng2016TPTcSrc}, and brute-force simulations on specialized hardware have resolved long-timescale protein dynamics and GPCR activation/binding mechanisms \cite{Shaw_Science10_Anton,Dror2011Beta2ARActivationMechanism,Dror2011GPCRDrugBindingPathway}. 
Finally, while alchemical FEP is well established for relative protein--ligand binding or closely related mutants, it becomes impractical when the alchemical changes induce large-scale conformational rearrangements that are slow and hard to overlap.

Recently, a complementary line of work has emerged from generative deep learning: \emph{Boltzmann generators and Boltzmann emulators} that aim to generate approximately independent configurations from an equilibrium distribution without integrating long MD trajectories. 
Boltzmann Generators are flow-based approaches that aim to generate independent samples from a Boltzmann distribution defined via an energy function \cite{NoeEtAl_19_BoltzmannGenerators,pmlr-v267-havens25a}. 
Boltzmann emulators approximate the equilibrium distribution
by training on MD simulation data and fine-tuning on experimental observables. This approach has recently become
popular for sampling protein ensembles, e.g., BioEmu \cite{bioemu2025,pmlr-v235-jing24a,janson2025asam,ZhengEtAl_NatMachIntell24_DiG,liu2025exendiff}.
The concept is also emerging in other application areas, such as material sciences \cite{ZhengEtAl_NatMachIntell24_DiG}. 
By producing effectively iid samples, these models address the \emph{slow-mixing} bottleneck.

However, iid sampling does not remove the \emph{rare-state} bottleneck: estimating observables controlled by low-probability regions still requires a number of samples that scales exponentially in free-energy differences. 
For example, BioEmu-1 estimates protein fold stabilities by directly sampling the equilibrium ensemble of protein structures and calculating the fraction of unfolded states \cite{bioemu2025}.
For stabilities of up to $\Delta G_{\mathrm{fold}}=-5$\,kcal/mol this is still feasible within one GPU hour, but this approach quickly becomes
intractable: sampling $\sim$1 in $1.9\times10^7$ unfolded samples for a protein with $\Delta G_{\mathrm{fold}}=-10$\,kcal/mol would take on the order of one GPU year \cite{bioemu2025}.


Here we provide a solution for the remaining sampling problems of diffusion-based equilibrium samplers
by integrating enhanced sampling models into the diffusion model framework. The key component is a steering algorithm that allows one to apply the desired bias potentials to a pretrained diffusion model at inference time. This enables the implementation of several classical enhanced sampling methods in the diffusion model framework, as well as the application of unbiasing methods such as WHAM \cite{FerrenbergSwendsen_PRL89_WHAM} or MBAR \cite{ShirtsChodera_JCP08_MBAR}. 
Furthermore, we present several technical improvements that provide low-variance estimates of the desired observables even if only a few thousand diffusion-model denoising trajectories can be afforded. Overall, this framework enables converged sampling of equilibrium properties of complex biomolecular processes with large energy differences within GPU minutes to hours per calculation, given a suitable pretrained diffusion model.
We demonstrate our framework using the BioEmu model on a biomolecular rare-event problem that is extremely difficult or currently impossible with all-atom MD simulations: efficient calculation of folding free energies for a variety of proteins ranging between 50 and 200 amino acids.

Related ideas of combining enhanced sampling methods with diffusion models have been presented in several recent publications: Nam et al.\cite{NamEtAl_EnhancingDMSampling} introduce a related approach employing adjoint sampling to fine-tune a pre-trained diffusion model towards the sampling of rare states. Richman and Dror\cite{RichmanDror_InferenceSteering} introduce a steering approach for exploring rare events with pretrained diffusion models with an approach similar to umbrella-sampling that can be viewed as special case of the present approach. Lam et al.\cite{LamEtAl_Metadiffusion} have presented a steering method which changes the sampled ensemble to be consistent with experimental data or enhance sample diversity.

\section{Biasing and unbiasing diffusion model ensembles}
\newcommand{\pfkt}{p_t^{FK}}
\newcommand{\defequal}{:=}
\newcommand{\Fbar}{\mathbb{E}_{x\sim q_t} F_t(x)}
\newcommand{\Fbatch}{F_t^\textrm{batch}}


We assume that we have a pretrained diffusion model whose output distribution for a given molecule of interest is $p(x)$ --- i.e., the model distribution $p(x)$ shall be our \textit{unbiased equilibrium distribution}. We will equivalently operate with probabilities and dimensionless energies, for example: 
\begin{equation}
    p(x) = \mathrm{e}^{-u(x)},
\end{equation}
with dimensionless energy $u(x)$. This representation is independent of the thermodynamic ensemble, e.g., in the canonical ensemble, $p(x)$ is the Boltzmann distribution and $u(x) = U(x) / k_B T$ with potential energy $U(x)$, Boltzmann constant $k_B$ and temperature $T$. Subsequently we will only assume that $p(x)$ and $u(x)$ exist, but not that they can be explicitly evaluated. Energy-based diffusion models, for which $p(x)$ and $u(x)$ can be efficiently evaluated, have been explored recently \cite{ArtsEtAl_2for1,plainer2025consistent}.

Our aim is to compute unbiased expectation values under the equilibrium distribution of the form
\begin{equation}
    O = \mathbb{E}_{p}[o(x)].
    \label{eq:equilibrium_expectation_value}    
\end{equation} 
For example, the probability of being in the unfolded state, $p_u$, can be cast in this form, but accurately estimating it---and thus the folding free energy, relies on sampling rare events. As in the traditional enhanced sampling literature, we proceed in two steps:
    


\begin{enumerate}
    \item \textbf{Generate biased ensembles}: given one or multiple biasing potentials $b_k(x)$, $k=1,...,K$, draw samples from the biased ensembles with energies $u(x) + b_k(x)$. This is achieved by steering the diffusion model at inference time (Section \ref{subsec:steering}).
    \item \textbf{Recover unbiased expectations}: reweight the biased samples given the known bias potentials in order to recover the unbiased expectation values. For $K=1$, reweighting is trivial, whereas for $K>1$ it can be performed with the multistate Bennett acceptance ratio (MBAR) method (Section \ref{subsec:weighted_MBAR}).
\end{enumerate}

\noindent
The choice of biasing potentials and strategies will be discussed in Sections \ref{sec:umbrelladiff}-\ref{sec:dGdiff}.


\subsection{Biased sampling from diffusion models}
\label{subsec:steering}

\paragraph{Unbiased diffusion model.} We assume our diffusion model has been trained to reverse the corruption process defined by a stochastic differential equation, \begin{equation}
\label{eq:forwardcorruption}
    dx_\tau = f_{1-\tau}(x_\tau)\,d\tau + \sigma_{1-\tau} dW_\tau, \qquad \textrm{$x_{\tau=0}\sim p_{\textrm{data}}$}
\end{equation}
Here $W$ is a standard Wiener process. The drift $f$ and diffusion $\sigma$ are chosen at training time such that $x\sim p_{\textrm{noise}} = N(0,I)$ at $\tau = 1$. We choose to write $f$ and $\sigma$ with $1-\tau$ as their argument rather than $\tau$ for notational convenience when writing about the reverse (denoising) process.
We parameterize reverse (denoising) time by $t \defequal 1-\tau$ and write
$p_t$ for the marginal density of $x_\tau$ at $\tau=1-t$. Thus, $p_0=p_{\textrm{noise}}$  and $p_1=p_{\textrm{data}}$.

 Training the diffusion model consists of learning to compute the score $\nabla \log p_t(x)$ as a function of $x$ and $t$.  
Sampling from the diffusion model is performed by first sampling $x$ from $p_{\textrm{noise}}$ and then simulating a reverse process of the form
\begin{align}\label{eq:unbiased_sde}
dx_t &= g_t(x_t)dt + \tilde{\sigma}_t dW_t\\
g_t(x) &:= -f_t(x) + \frac{\sigma_t^2 + \tilde{\sigma}_t^2}{2}\nabla \log p_t(x),
\end{align}
and $\tilde{\sigma}_t$ is a hyperparameter we can choose, with $\tilde{\sigma}_t=\sigma_t$ being the standard choice used in \cite{skreta2025feynman}.

\paragraph{Generating samples from tilted distribution.} Given a pretrained diffusion model which samples from $p(x)$, we seek a method to draw samples from a single biased ensemble defined via biasing potential $b(x)$: 
\begin{equation}
q(x)
\;\defequal\;
\frac{1}{Z}\,p(x)\,\mathrm{e}^{-b(x)},
\qquad
Z\defequal\int p(x)\,e^{-b(x)}\,\mathrm{d}x,
\label{eq:biased_ensemble}
\end{equation}
where $Z$ is an unknown normalization constant (partition function) that we will not need to evaluate explicitly.  

Several established families of steering methods exist:
(i) \emph{Score guidance}, which alters the reverse-time score by adding the gradient of a classifier, constraint, or reward to tilt generations toward desired properties \cite{DhariwalNichol2021Guidance,HoSalimans2022CFG,SongErmon2020SDE}.
(ii) \emph{Path-integral reweighting} 
via Sequential Monte Carlo (SMC) with Feynman-Kac (FK) potentials—keeping the proposal dynamics unchanged and accounting for the bias with incremental importance weights along the reverse trajectory; annealed importance sampling (AIS) appears as the no-resampling special case
\cite{neal2001ais,DelMoral2004SMC,DoucetJohansen2009SMC,Singhal2025FKSteering}.
(iii) \emph{Invariant correctors} (e.g., Metropolis-adjusted Langevin, MALA) that are interleaved with the reverse dynamics to mix within the current biased marginal without changing the FK weights \cite{RobertsTweedie1996MALA,DoucetJohansen2009SMC}.
Score-guidance methods are simple and fast but generally do not guarantee unbiased sampling of the target density. FK, AIS and SMC give explicit importance weights and, with resampling, can target the desired biased marginals exactly. Conceptually similar path-weighting ideas underlie the Hummer–Szabo reconstruction of equilibrium free-energy profiles from nonequilibrium pulling trajectories, where measurements collected along driven protocols are reweighted to recover unbiased thermodynamics \cite{HummerSzabo2001PNAS,HummerSzabo2010PNAS}.

Subsequently, we employ the Feynman-Kac Corrector (FKC) biased sampling methodology\cite{skreta2025feynman,ren2026driftlite}, described in the following paragraphs.

\paragraph{Biased marginals $q_t$.}
In order to sample from $q(x)$, we define a family of biased marginal densities $q_t$: \begin{equation}\label{eq:biased_marginals}
    q_t(x)\defequal \frac{p_t(x)e^{-b_t(x)}}{Z_t}, \qquad Z_t\defequal \int p_t(x)e^{-b_t(x)}dx
\end{equation}
where $b_0(x)$ is constant with respect to $x$, $b_1(x)\equiv b(x)$, and $b_t$ smoothly interpolates between the two. 


Skreta et al.\cite{skreta2025feynman} provide methods to modify the dynamics in Eq. \ref{eq:unbiased_sde} so that instead of sampling from $p_t$, we get weighted samples $(x_1,w_1), ..., (x_N, w_N)$ from biased marginals $q_t$. 
The weights $w_1, ..., w_N$ are importance weights, 
encoding the discrepancy between proposal and target biased marginals. 

In order to sample from the biased marginals in Eq. \ref{eq:biased_marginals}, we need a slightly more general version of the reward-tilted SDE of \cite{skreta2025feynman}. We get this by using control drift $\frac{\tilde{\sigma}_t^2}{2}\nabla b_t$ in the Driftlite framework\cite{ren2026driftlite}, resulting in the following system of SDEs:
\begin{align}
x_0 &\sim p_{\textrm{noise}} \label{eq:weighted_sde_1} \\
dx_t & = \Big(g_t(x) + \frac{\tilde{\sigma}_t^2}{2}\nabla b_t(x_t)\Big)dt + \tilde{\sigma}_t dW_t \label{eq:weighted_sde_2} \\
d \log w_t &= F_t(x_t)dt - \mathbb{E}_{x\sim q_t} [F_t(x)] \label{eq:weighted_sde_3} \\
F_t(x) &\defequal -\frac{\partial b_t(x)}{\partial t} +  \nabla b_t(x) \cdot  \left(f_t(x) - \frac{\sigma_t^2}{2}\nabla \log p_t(x)\right) \label{eq:weighted_sde_4}
\end{align}
If we choose $\sigma_t \equiv \tilde{\sigma}$ and $b_t(x)\equiv \lambda_t b(x)$ we recover the reward-tilted SDE\cite{skreta2025feynman}. 

\paragraph{Effective sample size}
We track the statistical efficiency of a weighted batch of $n$ samples using the Kish effective sample size (ESS) \cite{Kish65}:
\begin{equation}
    \mathrm{ESS} = \frac{\big(\sum_i w_i\big)^2}{\sum_i w_i^2}
    \label{eq:ESS}
\end{equation}
ESS ranges between 1 (all weight on one sample) and $n$ (all weights equal).
If the weights are normalized as $\sum_i w_i = 1$, it is $\mathrm{ESS} = 1 / \sum_i w_i^2$.

\paragraph{Resampling}
During denoising we periodically resample the particles using stratified sampling. Resampling can be done periodically, but should be done if the ESS falls below a predefined threshold, e.g., $n/2$ (Ref.~\cite{DoucetJohansen2009SMC}).
After each resampling step, the weights are refreshed, i.e.,
\begin{equation}
    (x_t^1, w_t^1), ..., (x_t^n,w_t^n) \rightarrow (\tilde{x}_t^1, 1), ..., (\tilde{x}_t^n,1).
\end{equation}
Optionally, we can enforce a resampling step after the steering procedure at $t=0$ per biased ensemble. In that case, all steered samples have uniform weights 1 and the subsequent expressions simplify.

\subsection{Unbiasing to the equilibrium distribution}
\label{subsec:weighted_MBAR}

After producing samples from $K$ biased ensembles (Sec.~\ref{subsec:steering}), we have a set of weighted samples $\{(x_{k,i}, w_{k,i})\}_{i=1}^{N_k}$ from each biased density $q_k$. Next we combine all biased samples into a single sample with weights $W_n$:
\begin{equation}
    \{(x_{k,i}, w_{k,i})\}_{i=1}^{N_k} \rightarrow \{(x_{n}, W_{n})\}_{n=1}^{N}.
    \label{eq:total_weight_unbiasing}
\end{equation}
The reweighting coefficients $W_n$ combine (i) the steering importance weights $w_{k,i}$ within each biased density $q_k$ produced via (\ref{eq:weighted_sde_1}-\ref{eq:weighted_sde_4}) and (ii) an additional unbiasing factor that maps each biased ensemble $q_k$ back to $p$.
If we have chosen to do a terminal resampling step at the end of steering, $w_{k,i} \equiv 1$ and $W_n$ only contains the unbiasing from $q_k$ to $p$. 

\paragraph{Equilibrium expectation values}
The weights $W_n$ must be such that we can express equilibrium expectation values of arbitrary observables $o(x)$ as:
\begin{equation}
\widehat{\mathbb{E}}_{p}[o] = \frac{\sum_{n=1}^{N} W_n\,o(x_n)}{\sum_{n=1}^{N} W_n}.
\label{eq:unbiased_expectation}
\end{equation}
in a way that is asymptotically unbiased in the infinite sample limit, i.e., $\lim_{N \rightarrow \infty} \widehat{\mathbb{E}}_{p}[o] = \mathbb{E}_{p}[o]$.
Expression \eqref{eq:unbiased_expectation} is sufficiently general that all quantities of the unbiased ensemble can be expressed with it. For example, to compute the folding free energy we define a membership function which assigns folded states to $1$ and unfolded states to $0$: 
\begin{equation}
    \chi(x) = \begin{cases}
        0 & x \ \,\mathrm{unfolded} \\
        1 & x \ \,\mathrm{folded}
    \end{cases},
\end{equation} 
$\chi(x)$ can itself be defined in terms of suitable order parameters, such as the fraction of native contacts from the folded structure, and it can optionally take intermediate values between 0 and 1. In terms of $\chi$, the folding free energy is defined by:
\begin{equation}
    \Delta G_\mathrm{fold} 
    = -k_B T \ln \frac{p_\mathrm{folded}}{p_\mathrm{unfolded}} 
    = -k_B T \ln \frac{\widehat{\mathbb{E}}_p[\chi]}{1-\widehat{\mathbb{E}}_p[\chi]}
\end{equation}
using \eqref{eq:unbiased_expectation}. Likewise, a potential of mean force (PMF) along a reaction coordinate $\xi(x)$ can be formulated by, e.g., binning the value range of $\xi$ and expressing the probability of each bin in terms of \eqref{eq:unbiased_expectation}. Practically, we obtain PMFs by constructing a weighted histogram or kernel density estimator from the weighted sample $\{(x_n, W_n)\}_{n=1}^{N}$.

\paragraph{Single biased ensemble}
For the special case $K=1$, i.e., using a single biased ensemble with bias potential $b(x)$, we have generated samples $\{(x_{n}, w_{n})\}_{n=1}^{N}$ (dropping the window index). The combined importance weights are then trivially given by direct reweighting:
\begin{equation}
    W_n = w_n \mathrm{e}^{b(x_n)}
    \label{eq:direct_reweighting}
\end{equation}

\paragraph{Unbiasing multiple biased ensembles with MBAR}
A statistically optimal method to combine samples from multiple biased ensembles towards unbiased ensemble estimators is the multistate Bennett acceptance ratio (MBAR), also known as binless weighted histogram analysis (WHAM) \cite{ShirtsChodera_JCP08_MBAR,KongEtAl_JRStatistSocB,Bartels_CPL00_MBAR}. For the special case $K=2$, MBAR becomes traditional BAR \cite{Bennett_JCP76_BAR}. 
MBAR yields minimum-variance, asymptotically unbiased estimates without requiring us to evaluate $\log p(x)$.
If the bias energy can be discretized we can instead work with histograms and binned WHAM \cite{FerrenbergSwendsen_PRL89_WHAM,Kumar1992,SouailleRoux_CPC01_WHAM}.

Here we describe a weighted MBAR variant which takes the weighted sample terminals from steering, $\{(x_{k,i},\,w_{k,i})\}_{i=1}^{N_k}$, and rewrites them into optimal reweighting coefficients for estimating expectations under $p$
(Eq. \ref{eq:total_weight_unbiasing}).
By convention, MBAR operates on $K+1$ ensembles, consisting of the unbiased ensemble $b_0(x) \equiv 0$ and the biased ensembles $b_k(x)$ for $k=1,\dots,K$. Note that we do not draw samples from ensemble 0, so sums run over sampled windows $\ell=1,...,K$.
To deal with weighted samples, we define the per-sample effective mass $\alpha_{k,i} \propto w_{k,i}$ and normalize so the window mass sums to the effective sample size of ensemble $k$, $\mathrm{ESS}_k$ (Eq. \ref{eq:ESS}):
\begin{equation}
    \alpha_{k,i} \;=\; \mathrm{ESS}_k \, \frac{w_{k,i}}{\sum_{j=1}^{N_k} w_{k,j}}
    \quad\Longrightarrow\quad
    \sum_{i=1}^{N_k} \alpha_{k,i} \;=\; \mathrm{ESS}_k .
\end{equation}
If we have chosen to do terminal resampling at the end of steering, the input samples are already unweighted, $w_{k,i} \equiv 1$, $\mathrm{ESS}_k=N_k$,
hence $\alpha_{k,i}=1$ and we recover standard MBAR.

We then aggregate all samples as $\{x_n\}_{n=1}^N$ with associated
$\alpha_n$ (from the bias potential with which they were generated). Let
$M_\ell = \sum_{n:\,\mathrm{orig}(x_n)=\ell} \alpha_n$ denote the window mass, where
$M_\ell=\mathrm{ESS}_\ell$ under the default above, and $\mathrm{orig}(x_n)$ denotes the index of the window in which sample $x_n$ was generated.
We solve for the reduced free energies of the sampled biased states 
$k=1,...,K$ via
\begin{equation}
\exp(-\hat f_k)
\;=\;
\sum_{n=1}^{N}
\frac{\alpha_n\,\exp\!\big(-b_k(x_n)\big)}
{\sum_{\ell=1}^{K} M_\ell \exp\!\big(\hat f_\ell - b_\ell(x_n)\big)},
\label{eq:wmbar_fixed_point}
\end{equation}
and forms target-state weights ($a\in\{0,1,\dots,K\}$), with the unbiased state $a=0$
included as the target for reweighting, as
\begin{equation}
W_n^{(a)}
\;=\;
\frac{\alpha_n\,\exp\!\big(-b_a(x_n)\big)}
{\sum_{\ell=1}^{K} M_\ell \exp\!\big(\hat f_\ell - b_\ell(x_n)\big)}.
\label{eq:wmbar_weights}
\end{equation}
We then compute unbiased expectations in \eqref{eq:unbiased_expectation} using
\begin{equation}
W_n \equiv W_n^{(0)}.
\label{eq:wmbar_expectation}
\end{equation}

Using $\alpha_{k,i}\propto w_{k,i}$ with the normalization above lets MBAR
automatically down-weight windows with highly skewed particle weights via their
$\mathrm{ESS}_k$. If desired, one can replace $\mathrm{ESS}_k$ by an external
\emph{effective count} (e.g.\ number of denoising trajectories used for window $k$,
or a correlation-corrected $N_k/g_k$); equations
\eqref{eq:wmbar_fixed_point}-\eqref{eq:wmbar_expectation} are unchanged.

To compute uncertainties, we can use MBAR's covariance expressions \cite{ShirtsChodera_JCP08_MBAR},
or apply a cluster bootstrap that resamples at the level of independent denoising
trajectories to account for within-window correlations, e.g., from late branching.


\subsection{Illustration of enhanced diffusion sampling}
\begin{figure*}
    \centering    
    \textbf{a)}\includegraphics[width=0.2\linewidth]{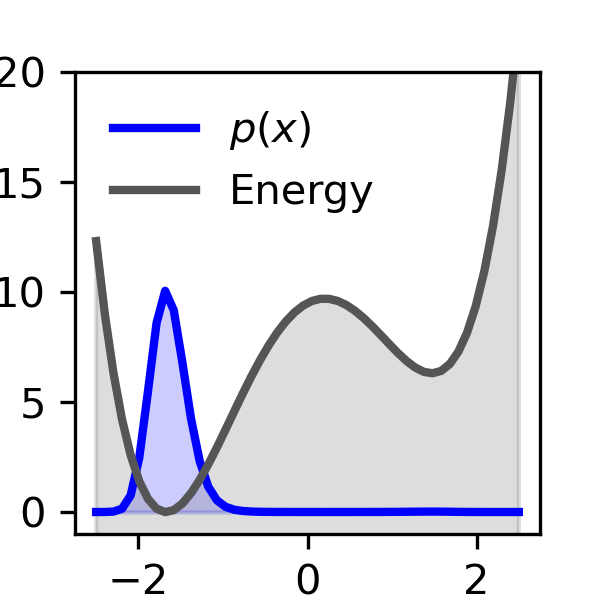}
    \textbf{b)}\includegraphics[width=0.2\linewidth]{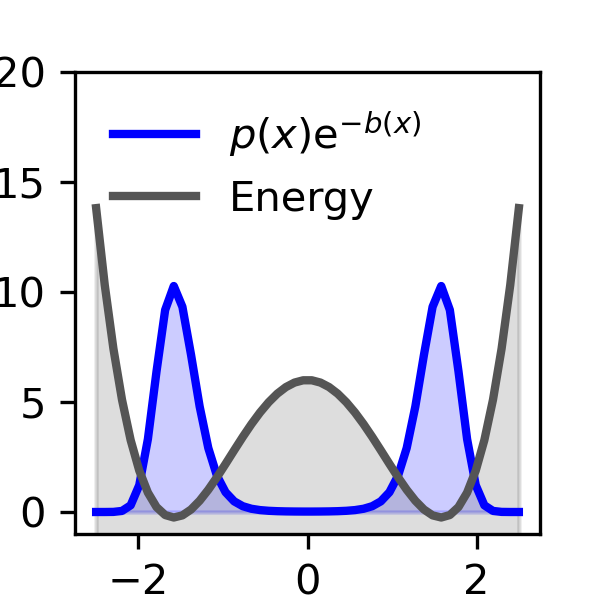}
    \textbf{d)}\includegraphics[width=0.5\linewidth]{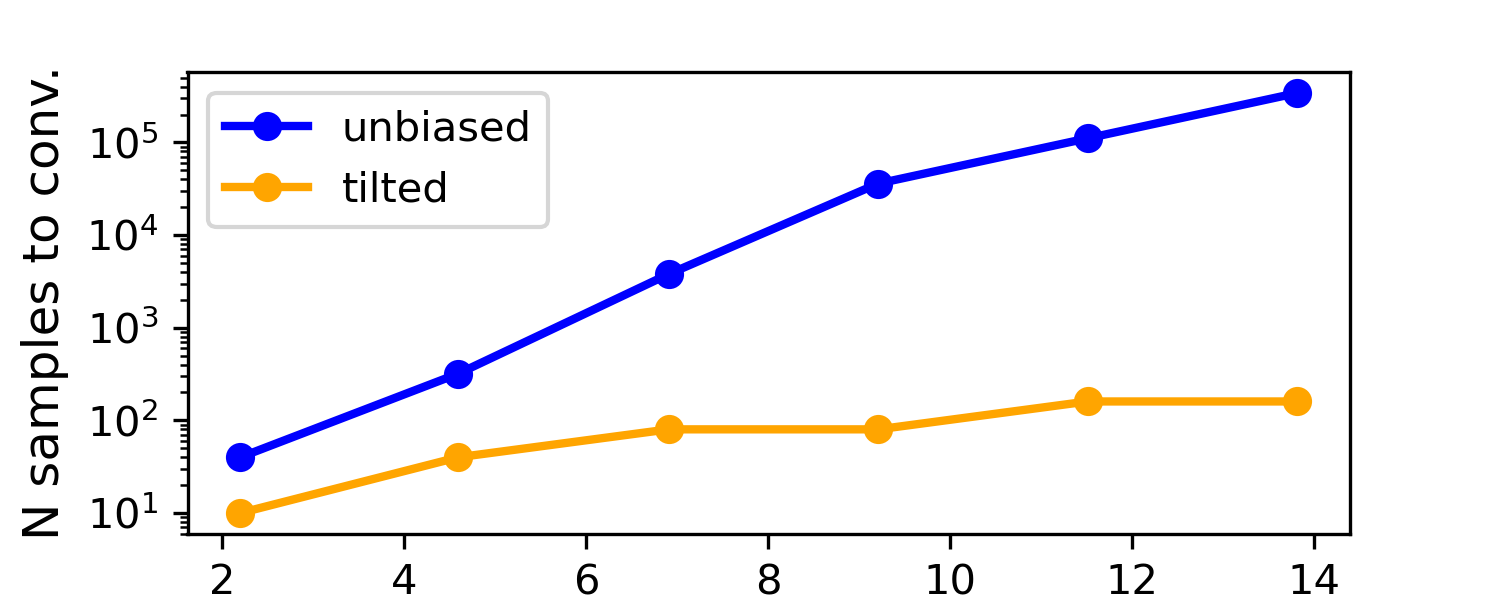} \\
    \textbf{c)}\includegraphics[width=0.7\linewidth]{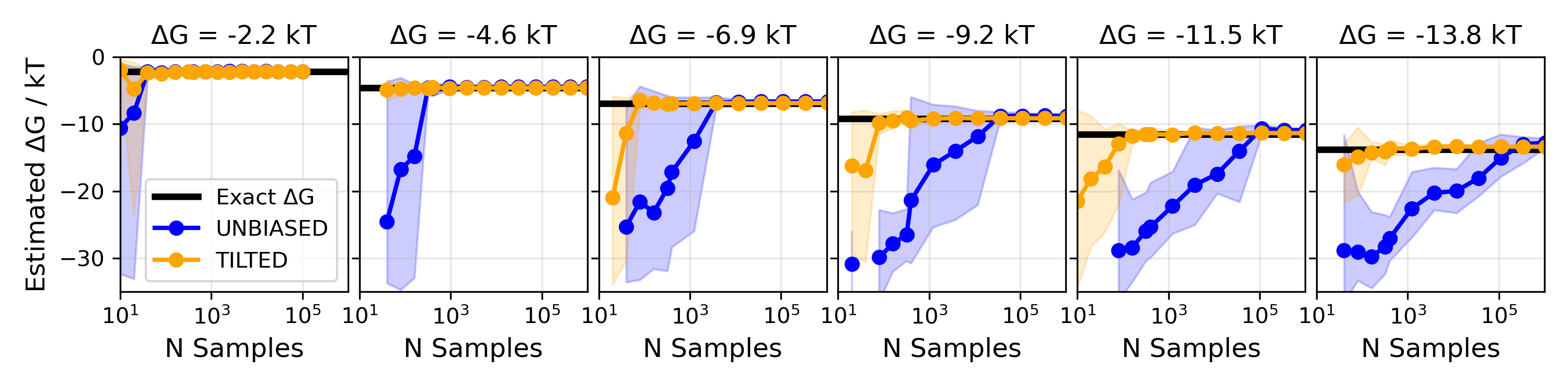}
    \caption{\textbf{Enhanced sampling with diffusion models efficiently computes probabilities of rare events} 
    \textbf{a)} Unbiased diffusion model probability density $p(x)$ and corresponding energy $-\ln p(x)$. In this example the free energy difference between the two minima is set to $\Delta G = -7\,k_BT$.
    \textbf{b)} Biased density $p(x) \mathrm{e}^{-b(x)}$ using a linear tilt $b(x) = \beta x$ with $\beta$ chosen to achieve $\Delta G_{\mathrm{biased}} = 0\,k_BT$.
    \textbf{c)} Estimates of $\Delta G$ with an unbiased diffusion model and enhanced sampling using a tilting potential that sets $\Delta G_{\mathrm{biased}} = 0\,k_BT$. Mean (solid line) and 95\% confidence interval (shaded area) are shown.
    \textbf{d)} Number of samples required to get an estimate within 1 kcal/mol of the exact $\Delta G$ value and with a standard deviation of 1 kcal/mol, as a function of $\Delta G$.}
    \label{fig:1d_tilting}
\end{figure*}

Before moving on to more complex enhanced
sampling protocols, we first demonstrate
that enhanced diffusion sampling is efficient in an idealized setting (Fig.~\ref{fig:1d_tilting}).
We define a family of double-well potentials
$u(x)$ where the energy difference between the two minima is given by $\Delta G$ ranging from -2 to -14 $k_BT$. The corresponding equilibrium densities are given by $p(x) \propto \mathrm{e}^{-u(x)}$,
resulting in equilibrium probabilities between $10^{-1}$ and $10^{-6}$ for the high-energy state (Fig.~\ref{fig:1d_tilting}a).
For each $\Delta G$ value, a diffusion model with an analytical score function is defined that generates samples from $p(x)$ exactly.
To evaluate the effect of enhanced diffusion sampling on sampling efficiency directly, we define a single linear bias potential for each double-well potential, $b_{\Delta G} = a(\Delta G)\,x$ where $a(\Delta G)$ is chosen such that the two energy minima are equal in the biased potential (Fig.~\ref{fig:1d_tilting}b). 

For each $\Delta G$ value we then generate $N$ samples with each of two protocols: (i) direct sampling from the unsteered diffusion model with density $p(x)$, and (ii) steered sampling as described in Sec. \ref{subsec:steering} using $b_{\Delta G}$ and direct reweighting of the biased ensemble with \eqref{eq:direct_reweighting}. We then estimate $\Delta G$ from the samples and compare with the exact value.
Both protocols converge to the correct $\Delta G$ value in the large $N$ limit (Fig.~\ref{fig:1d_tilting}c).
Enhanced diffusion sampling converges faster than equilibrium sampling for all inspected $\Delta G$ values and the performance gap increases with $\Delta G$. 
To compare sampling efficiencies between unbiased and enhanced sampling protocols, we run five independent repeats for each $N$ and each protocol and define a sample as being converged when both the absolute difference of the empirical mean of $\Delta G$ from the exact value, and the empirical standard deviation of $\Delta G$ are within 1 kcal/mol (at 300K) across repetitions.
As expected, the number of samples needed to reach convergence with unbiased sampling increases approximately exponentially with $\Delta G$ (Fig.~\ref{fig:1d_tilting}d, blue). In contrast, the number of samples required by enhanced diffusion sampling only increases mildly with $\Delta G$ and ranges between 10 to 100 samples for all $\Delta G$ values inspected here (Fig.~\ref{fig:1d_tilting}d, orange).

We note that this setup describes the best-case scenario in which only a single biasing potential is needed and the perfect bias is known in advance. In a realistic scenario we will have to try different biasing potentials until all desired states have been sampled, and then combine them with MBAR, which will lead to a greater number of samples needed to convergence and will complicate the relationship between free energy differences and the number of samples needed.

\section{UmbrellaDiff --- umbrella sampling with diffusion models}
\label{sec:umbrelladiff}

Here we describe how the classical umbrella sampling method \cite{Torrie_JCompPhys23_187} can be adapted to steered diffusion models. The goal of umbrella sampling is to
compute the marginal probability density $p_\Xi(\xi) = \int p(x) \delta(\xi(x)-\xi) dx$, or free energy profile---also known as potential of mean force (PMF),
\begin{equation}
    u_\Xi(\xi) = -\ln p_\Xi(\xi), \qquad U_\Xi(\xi) = k_B T u_\Xi(\xi),
    \label{eq:PMF_1d}
\end{equation}
along a user-chosen reaction coordinate
$\xi:\mathcal{X}\to\mathbb{R}$, e.g., the end-to-end distance of a protein, or the fraction of native contacts.
We will formulate umbrella sampling for the simplest case of one-dimensional coordinates, but the extension to multidimensional umbrella sampling is straightforward \cite{SouailleRoux_CPC01_WHAM}.

Umbrella sampling introduces a set of $K$ biased ensembles, each of which restrains sampling near prescribed centers $c_k$ in $\xi$-space. The most common bias potential is harmonic with stiffness $\kappa_k>0$:
\begin{equation}
b_k(x)=\tfrac{1}{2}\,\kappa_k\left\Vert\xi(x)-c_k\right\Vert^2,
\label{eq:umbrella_bias}
\end{equation}
although other choices are possible, e.g., flat-bottom with harmonic tails.
For each bias potential, we generate an ensemble using the steering procedure of Sec.~\ref{subsec:steering} and remove bias with weighted MBAR (Sec.~\ref{subsec:weighted_MBAR}). MBAR returns a set of equilibrium weights for each sample, $\{(x_n, W_n)\}$, which can then be used in a histogram or kernel density estimate to obtain $p_\Xi(\xi)$. Finally we can obtain the PMF via Eq. (\ref{eq:PMF_1d}).
In order to obtain smooth estimates of $p_\Xi(\xi)$ it can be beneficial to increase the number of samples using late branching near $t\!\approx\!0$ (Sec.~\ref{subsec:steering}).

\paragraph{Designing bias potentials to ensure coverage and overlap.}
We choose centers $\{c_k\}_{k=1}^K$ to cover the $\xi$-range of interest. Our prior model of the biased density is that induced by the bias potential alone, i.e., a Gaussian with center $c_k$ and standard deviation
$\sigma_k \approx 1/\sqrt{\kappa_k}$. 
As a rule of thumb, space neighboring centers within one or two $\sigma$ to obtain
$\sim 10$-$30\%$ overlap along $\xi$. 

As the effect of $p(x)$ will change the resulting distribution of $q_k(x)$, the bias parameters, may be chosen adaptively, e.g., with a preconditioning such as: (i) start with evenly spaced $c_k$ and 
$\kappa_k = 1/(c_k-c_{k-1})^2$,
(ii) run a small initial sample and estimate empirical $\mathrm{Var}_{q_k}[\xi]$, (iii) update 
$\kappa_k \leftarrow 1/\widehat{\mathrm{Var}}_{q_k}[\xi]$ 
to equalize widths.

\paragraph{Diagnostics}
We recommend to monitor standard diagnostics from umbrella sampling / free-energy analysis to assess (a) whether each biased ensemble is well sampled and (b) whether the set of windows forms a connected path that permits low-variance reweighting
\cite{Kastner2011UmbrellaSampling,Henin_Lelievre_Shirts_Valsson_Delemotte_2022,Klimovich2015Guidelines}.
In particular the following diagnostics are suitable:
\begin{enumerate}
    \item Per-window effective sample size (ESS) of the steering terminal weights
    $w_{k,i}$ (Eq.~\ref{eq:ESS}); when terminals may be correlated (e.g.\ due to late branching or MCMC rejuvenation), we additionally report a correlation-corrected effective count $N^{\mathrm{eff}}_k \approx N_k/g_k$ using the statistical inefficiency $g_k$ and use a cluster/bootstrap analysis for uncertainties \cite{ChoderaEtAl_JCP11_RateTheory}.
    \item Window overlap. Along the umbrella coordinate $\xi$, we quantify overlap between neighboring windows using histogram/KDE-based overlap coefficients (e.g.\ $O_{k,k+1} = \int \sqrt{q_k(\xi) q_{k+1}(\xi)} d\xi$ or related normalized measures).
    More generally, we compute the \emph{state overlap matrix} $O_{ij}$ from MBAR/WHAM, where $O_{ij}$ estimates the probability that a configuration sampled from state $j$ could be observed in state $i$. 
    As a practical rule, require neighboring-window overlap area in the range 0.1 to 0.3 and a well-connected overlap matrix (no near-block-diagonal structure) by inserting additional windows or revising bias parameters \cite{Klimovich2015Guidelines,ShirtsChodera_JCP08_MBAR,pymbar2022}
    \item MBAR/WHAM uncertainty estimates, including the asymptotic covariance/standard errors for free energies and target observables \cite{ShirtsChodera_JCP08_MBAR,Kumar1992}, and stability of the PMF/observables under increasing sample size (and optionally leave-one-window-out checks). 
    For umbrella sampling in particular, we optionally complement MBAR with EMUS error analysis to identifying windows that dominate variance and guide adaptive allocation of sampling effort \cite{Thiede2016EMUS}.
\end{enumerate}
These diagnostics can be used to adapt window centers and force constants, increase sampling in selected windows, or insert bridging windows to restore overlap \cite{Kastner2011UmbrellaSampling,Klimovich2015Guidelines}.

\paragraph{UmbrellaDiff versus traditional umbrella sampling with MD}
Compared to traditional umbrella sampling with MD, UmbrellaDiff has several advantages. First, in conventional umbrella sampling one must prepare initial configurations for each window (often via pulling/steered protocols). Then one must equilibrate sufficiently so that neighboring windows overlap not only in the restrained coordinate $\xi$ but also in orthogonal, slow degrees of freedom \cite{Kastner2011UmbrellaSampling,Klimovich2015Guidelines,DasVenkatramani2025AutoSIM}.
This is not an issue with UmbrellaDiff, as each steered sample is independently generated from the noisy state using the procedure in Sec. \ref{subsec:steering}.

\begin{figure*}
    \centering    \includegraphics[width=0.8\linewidth]{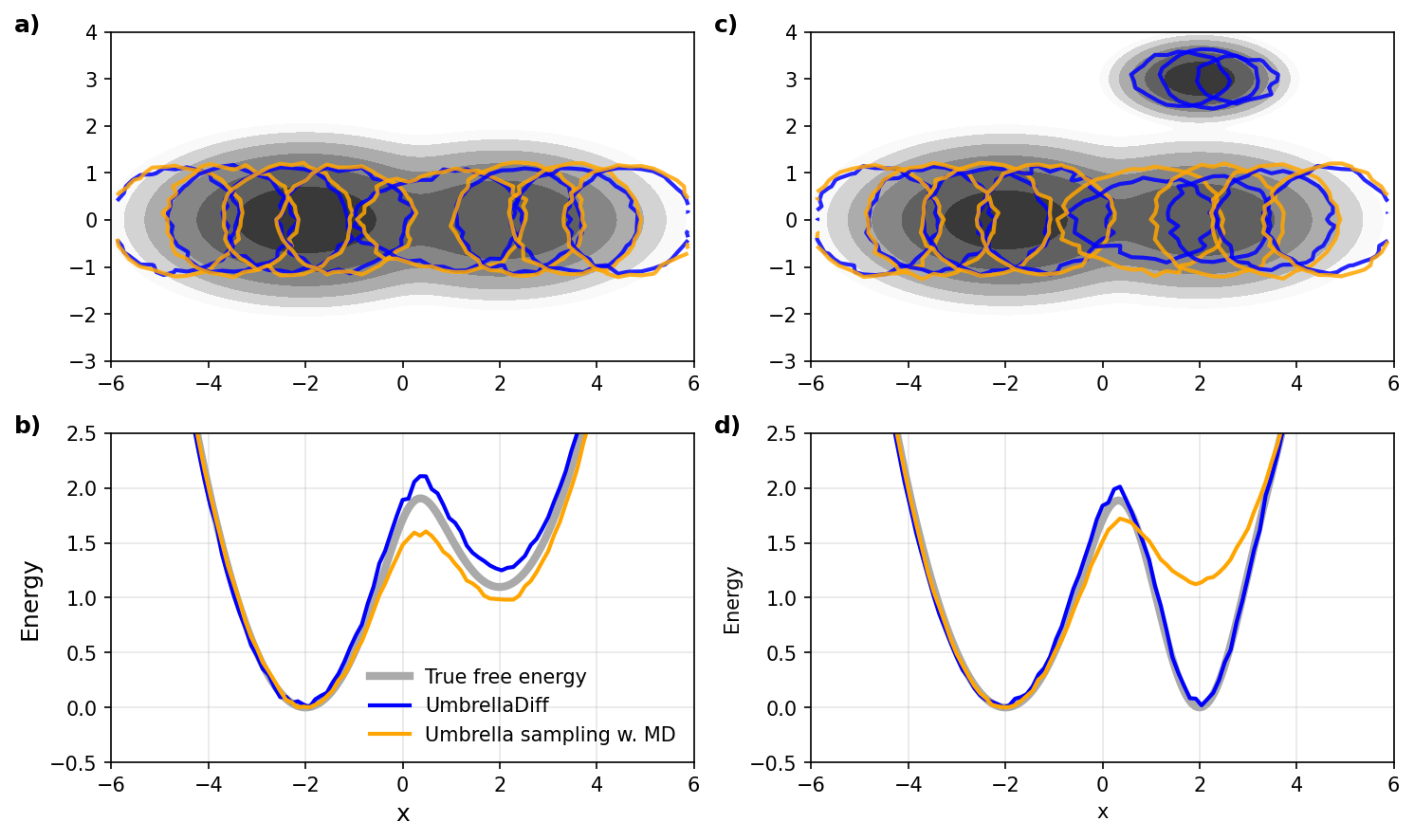}
    \caption{
    \textbf{UmbrellaDiff bypasses kinetic trap problem in traditional umbrella sampling.}
    a) Simple two-state 2D potential. Umbrella sampling is performed using 8 umbrellas that bias $\xi = x$ using regular umbrella sampling with Langevin dynamics (orange) and UmbrellaDiff (blue). Contours show 95\% of the sample density of individual umbrellas.
    b) Free energy as a function of $x$ using traditional umbrella sampling (orange) and UmbrellaDiff (blue). Both methods approximate the true potential of mean force (grey).
    c) As a) but with a third high-probability state (top right). Traditional umbrella sampling (orange) does not sample the third state as it is off-path and long simulation trajectories in each umbrella would be required to sample this state. UmbrellaDiff (blue) samples the equilibrium distribution conditioned on the umbrella potential and therefore samples the third state without issues.
    d) As b). Traditional umbrella sampling estimates a wrong free energy due to failing to sample the third state (orange), UmbrellaDiff estimates the correct free energy (blue).
    }
    \label{fig:pmf_two_paths}
\end{figure*}

Second, when hidden barriers or multiple channels exist in directions perpendicular to $\xi$, traditional umbrella simulations can remain trapped in distinct metastable “orthogonal states”, leading to hysteresis and very slow convergence of the PMF unless additional sampling machinery is introduced (Fig.~\ref{fig:pmf_two_paths}) \cite{Yang2014ITSUS,Sousa2023STeUS}.
This is not an issue for UmbrellaDiff, where the only role of the umbrellas is to ensure that low probability values in $p_\Xi(\xi)$ are sampled, while the iid sampling property of the diffusion model avoids problems with kinetic trapping and slow intra-window relaxation. Fig.~\ref{fig:pmf_two_paths} demonstrates that UmbrellaDiff can compute unbiased free energy profiles even if metastable states that are off of the main path of umbrellas in $\xi$ contribute significantly to the free energy, whereas traditional umbrella sampling struggles with these cases.

\section{MetaDiff --- metadynamics with diffusion models}
\label{sec:metadiff}

Metadynamics \cite{LaioParrinello_PNAS99_12562,LaioGervasio_RPP08_Metadynamics,BarducciBonomiParrinello_WIREs11} is a widely used enhanced sampling method with many follow-up variants, including well-tempered metadynamics \cite{BarducciBussiParrinello_PRL08_WellTemperedMetadynamics,DamaParrinelloVoth_PRL14} and associated estimators \cite{TiwaryParrinello_JPCB15}.
The original idea of metadynamics is to run an MD simulation and to periodically deposit repulsive \textit{hills} (typically Gaussians) in the space of chosen collective variables (CVs) $\xi(x)$ at the current CV value. The resulting history-dependent bias progressively fills free-energy minima and drives exploration of new regions \cite{LaioParrinello_PNAS99_12562,LaioGervasio_RPP08_Metadynamics}.
In standard metadynamics, the bias keeps growing, while in well-tempered metadynamics the hill heights are tempered so that the bias converges (up to a known factor) to the negative free energy \cite{BarducciBussiParrinello_PRL08_WellTemperedMetadynamics,DamaParrinelloVoth_PRL14}. 
Because the bias is explicitly time-dependent, metadynamics is generally analyzed via the accumulated bias and dedicated reweighting/estimators rather than by treating the trajectory as equilibrium sampling at a fixed thermodynamic state; in particular, during the initial ``fill-up'' period it is a nonequilibrium process, so equilibrium multi-state estimators such as MBAR are not directly applicable \cite{LaioGervasio_RPP08_Metadynamics,TiwaryParrinello_JPCB15}.

Here we formulate a metadynamics variant for steered diffusion models. In contrast to biased MD trajectories, each invocation of the steering algorithm (Sec.~\ref{subsec:steering}) targets the equilibrium distribution of the \emph{current} bias condition and returns iid (weighted) samples from that biased ensemble. Thus, each bias update defines a well-posed thermodynamic state, and we \emph{can} apply MBAR online to combine samples across bias conditions and obtain an unbiased free-energy estimate (and diagnostics) at any time, without waiting for a fully ``filled'' surface. 
Finally, since diffusion inference is naturally batched, MetaDiff updates the bias from a batch of samples: instead of a single Gaussian hill, we add a batchwise kernel (a mixture of Gaussians in $\xi$-space). In the limit of batch size 1, the procedure reduces to standard metadynamics applied to steered diffusion sampling.

We will again formulate the method on a one-dimensional reaction coordinate
$\xi:\mathcal{X}\to\mathbb{R}$, the generalization to multiple dimensions is conceptually straightforward. 
We represent the bias at outer iteration $k$ as a sum of Gaussians:
\begin{align}
b_k(\xi)&\;=\;\sum_{m=1}^{M_k} a_m\,K_\sigma(\xi;\mu_m), \\
K_\sigma(\xi;\mu)&\;=\;\frac{1}{\sqrt{2\pi}\,\sigma}\exp\!\Big(-\frac{(\xi-\mu)^2}{2\sigma^2}\Big),
\end{align}
with $b_0\equiv0$. 
For each outer iteration $k$, we produce a batch of samples with normalized weights using the steering algorithm (Sec.~\ref{subsec:steering}):
$\{(x_{k,i}, w_{k,i})\}_{i=1}^{N_k}$ with $\xi_{k,i}=\xi(x_{k,i})$. 
We then deposit one Gaussian per sample sequentially so that $b_k$ remains a sum of Gaussians at all times, in the spirit of classical metadynamics \cite{LaioParrinello_PNAS99_12562} and well-tempered metadynamics \cite{BarducciBussiParrinello_PRL08_WellTemperedMetadynamics}. To control the total amount of bias added at iteration $k$, we choose a per-iteration mass $h_k>0$ and split it across the $N_k$ samples using per-Gaussian masses $\tilde h_{k,i}$ that satisfy $\sum_{i=1}^{N_k}\tilde h_{k,i}=h_k$. A simple choice is
\begin{equation}
    \tilde h_{k,i}=\begin{cases}
h_k/N_k, & \text{for unweighted samples},\\[4pt]
h_k\, \frac{w_{k,i}}{\sum_j w_{k,j}}, & \text{for importance-weighted samples},
\end{cases}
\label{eq:h_ki}
\end{equation}
so each iteration deposits total mass $h_k$ regardless of batch size. In practice, standard metadynamics often uses a \emph{constant} base height ($h_k\equiv h_0$) and relies on exploration to distribute hills, whereas well-tempered metadynamics also works robustly with a constant $h_k$ because the effective deposited height decays automatically via tempering \cite{BarducciBussiParrinello_PRL08_WellTemperedMetadynamics}; optionally, a gently decaying schedule $h_k=h_0\rho^k$ with $\rho\in(0,1)$ reduces the noise during late iterations.

The final amplitude for each Gaussian $i$ is determined by whether we use a standard or well-tempered Metadynamics scheme:
\begin{align}
\text{(standard)}\qquad
& a_{k,i}\;=\;\tilde h_{k,i}, 
\label{eq:std_stream_amp}\\[2pt]
\text{(well-tempered, $\gamma>1$)}\qquad
& a_{k,i}\;=\;\tilde h_{k,i}\,\exp\!\Big(-\frac{b_k^{(i-1)}(\xi_{k,i})}{\gamma-1}\Big).
\label{eq:wt_stream_amp}
\end{align}
Setting $\gamma=\infty$ in
\eqref{eq:wt_stream_amp} recovers the standard rule \eqref{eq:std_stream_amp}.
Finally, we update the bias. Starting from $b_k^{(0)}\equiv b_k$ we sequentially iterate through the batch for $i=1,\dots,N_k$:
\begin{equation}
b_k^{(i)}(\xi)\;=\;b_k^{(i-1)}(\xi)\;+\;a_{k,i}\,K_\sigma(\xi;\xi_{k,i}),
\qquad \xi_{k,i}=\xi(x_{k,i}).
\label{eq:seq_update}
\end{equation}
After processing all $N_k$ terminals, we set $b_{k+1}\equiv b_k^{(N_k)}$ as the new bias potential.

\paragraph{Algorithm (\textsc{MetaDiff}).}
Given kernel width $\sigma$, tempering factor $\gamma\ge 1$, and a base-height schedule $\{h_k\}_{k\ge0}$:
\begin{enumerate}
\item Initialize $b_0\equiv0$.
\item For $k=0,1,2,\dots$:
  \begin{enumerate}
  \item Sample from $q_{k}$ via Steering (Sec. \ref{subsec:steering}) and collect weighted samples $\{(x_{k,i}, w_{k,i})\}_{i=1}^{N_k}$. 
  \item Choose $\tilde h_{k,i}$ using Eq. (\ref{eq:h_ki}).
  \item For $i=1,\dots,N_k$: compute $a_{k,i}$ by \eqref{eq:std_stream_amp} (standard) or
        \eqref{eq:wt_stream_amp} (well-tempered) using the current $b_k^{(i-1)}$, then update $b_k^{(i)}$ via \eqref{eq:seq_update}.
  \item Set $b_{k+1}\leftarrow b_k^{(N_k)}$. Stop when $\sup_\xi|b_{k+1}(\xi)-b_k(\xi)|<\varepsilon_b$
        or when an MBAR-based PMF change across selected checkpoints falls below $\varepsilon_F$.
  \end{enumerate}
\end{enumerate}

\begin{figure*}[ht]
    \centering
    \includegraphics[width=0.8\linewidth,trim=0 50 0 10]{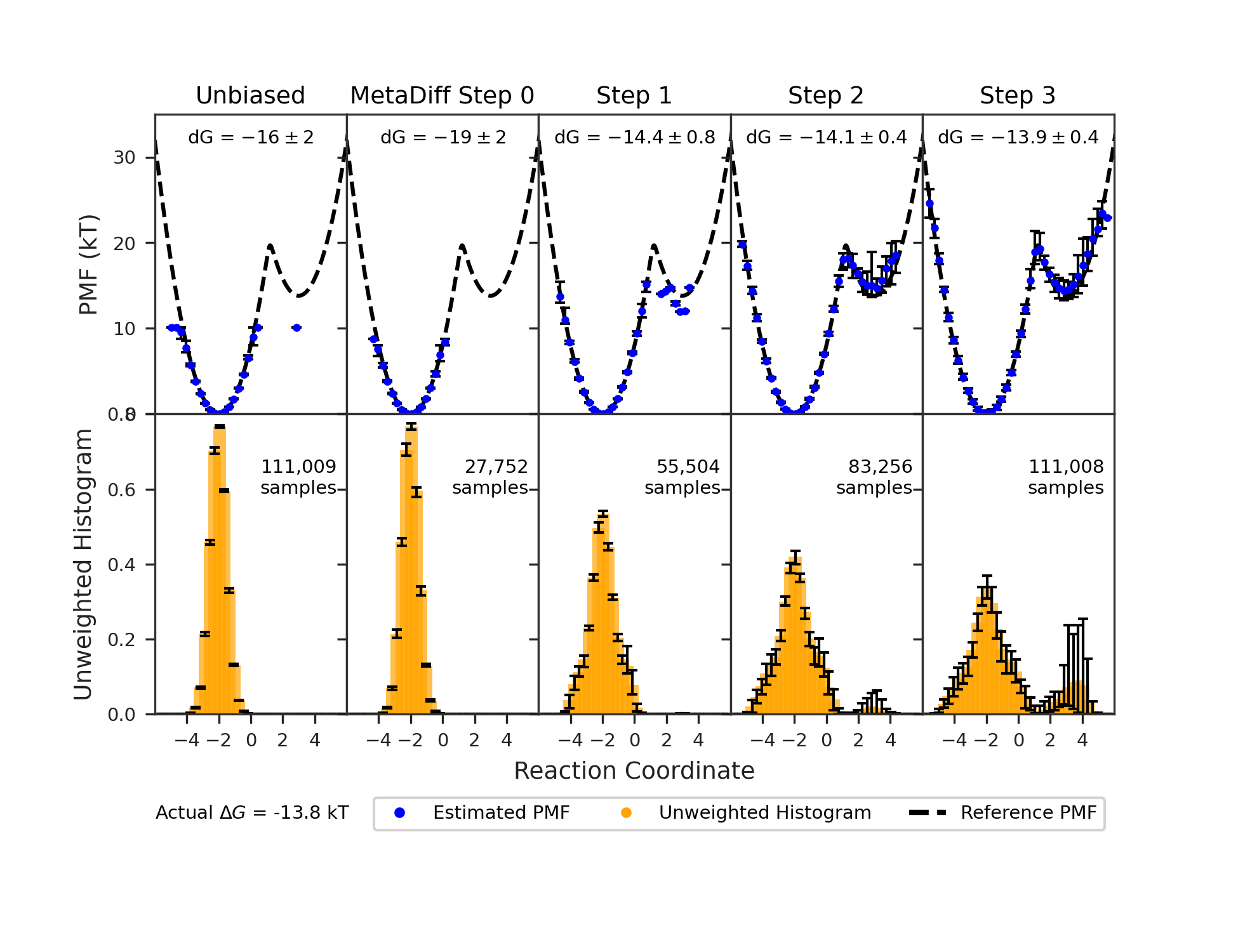}
    \caption{MetaDiff illustration on 1-dimensional double-well potential. Estimated PMF using iterations of metadynamics at a free energy difference of $\Delta G=-13.8 k_BT$.}
    \label{fig:metadiff_1d}
\end{figure*}

\paragraph{Mean-field consistency and stationary PMF.}
As $\sigma\!\to\!0$ and $h_k\!\to\!0$ (small hills), the expected increment at $\xi$ under
\eqref{eq:wt_stream_amp}-\eqref{eq:seq_update} satisfies
\[
\mathbb{E}\big[b_{k+1}(\xi)-b_k(\xi)\big]\;\propto\; q_{\Xi,k}(\xi)\,
e^{-b_k(\xi)/(\gamma-1)},
\]
which is the well-tempered mean-field update \cite{BarducciBussiParrinello_PRL08_WellTemperedMetadynamics}. At stationarity,
\(
u_{\Xi}(\xi)=-(\gamma/(\gamma-1))\,b_\infty(\xi)+C
\)
(with arbitrary constant $C$); 
For $\gamma=\infty$ the update reduces to standard metadynamics; in the idealized fully-filled limit the accumulated bias approximates 
$-u_{\Xi}(\xi) + C$
\cite{LaioParrinello_PNAS99_12562}.
The typical practice is to keep $h_k$ constant and choose $\gamma$ to tune smoothness/convergence
($\gamma\approx 5$-$20$ is common). A modest decay of $h_k$ can further stabilize late iterations.
\paragraph{Illustration on a double-well potential}
Fig.~\ref{fig:metadiff_1d} demonstrates MetaDiff on a one-dimensional double-well potential whose energy minima have a free energy difference of -13.8 $k_BT$. Using $1.11 \cdot 10^5$ samples, the unbiased diffusion sampler only samples the lower energy minimum but fails to sample the second minimum (Fig.~\ref{fig:metadiff_1d}, first column). 
MetaDiff is run in batches of $1.77 \cdot 10^4$ samples. The second minimum is explored after the second batch and an accurate estimate of the free energy difference is immediately achieved, and further improved in subsequent batches (Fig.~\ref{fig:metadiff_1d}, columns 2-5).

\section{$\Delta G$-Diff: Free energy differences between two states}
\label{sec:dGdiff}

Here we aim to compute the free energy difference between two configurational states $A$ and $B$, $\Delta G_{\mathrm{AB}}$. Examples include a folding free energy, binding free energy, the free energy change of a reaction or a conformational change.
Let $\xi(x)\in[0,1]$ be a smooth progress coordinate that indicates how far a configuration $x$ has advanced from state $A$ to $B$:
\begin{equation}
    \xi(x) = \left\{\begin{array}{lr}
        0 & x \in A \\
        (0,1) & x \notin \{A,B\} \\
        1 & x \in B \\
    \end{array} \right.
\end{equation}
In practice, $\xi$ can be a sigmoid function of a differentiable fraction-of-native-contacts (FNC), a normalized root-mean-squared-deviation (RMSD) or a committor function defined between sets of configurations $A$ and $B$ \cite{Onsager_PhysRev1938_SplittingProbability,BolhuisDellagoChandler_PNAS00_Ala2ReactionCoordinates,EVandenEijnden_TPT_JStatPhys06,PetersTrout_JCP06_ReactionCoordinateOptimization}. 

To ensure that both states can be sampled, we bias with a centered linear tilt in reduced units
\[
b_a(x)=a\,\Big(\xi(x)-\tfrac12\Big),\qquad a\in\mathbb{R},
\]
so $a>0$ favors $A$ ($\xi\!\approx\!0$) and $a<0$ favors $B$ ($\xi\!\approx\!1$).

\paragraph{Dominance and overlap diagnostics.}
We define core sets in the $\xi$ coordinate as $C_A=[0,c]$, $C_B=[1-c,1]$ with, e.g., $c=0.4$.
From a weighted batch with DM steering (Sec. \ref{subsec:steering}) with tilt $a$ we have samples and normalized weights ${(\xi(x_i), \bar w_i)}$. We define the core probability masses as
\begin{align}
\widehat{\pi}_A(a)&=\sum_i \bar w_i\,\mathbf{1}\{\xi(x_i)\in C_A\} \\
\widehat{\pi}_B(a)&=\sum_i \bar w_i\,\mathbf{1}\{\xi(x_i)\in C_B\}.
\end{align}
Say $a$ has \emph{$A$-dominance} if $\widehat{\pi}_A(a)\ge \vartheta$ and \emph{$B$-dominance} if $\widehat{\pi}_B(a)\ge \vartheta$ with default $\vartheta=0.5$ (can be increased on demand).

For adjacent tilts $a<a'$, we measure the overlap as follows
\[
\widehat{\mathcal{O}}_{a\to a'} \;\propto\; \sum_{i\in a}
\exp\!\big(\!-\,(a'-a)\,[\xi(x_i)-\tfrac12]\big),
\]
and require $\widehat{\mathcal{O}}_{a\to a'},\,\widehat{\mathcal{O}}_{a'\to a}\ge \varepsilon_{\rm ov}$ (e.g.\ $10$-$30\%$).

\paragraph{Algorithm ($\Delta$G-Diff).}
Inputs: tilt step $s>0$ (default\ $s\approx 2$ in $k_BT$), initial sample size $n$ (default 100), dominance threshold $\vartheta$ (default 0.5), overlap threshold $\varepsilon_{\rm ov}$ (default 0.25), ESS$_\mathrm{target}$ (default 100).
\begin{enumerate}
\item \textbf{Initialize:} $\mathcal{A}\leftarrow\{0\}$. Sample $n$ terminals with $a{=}0$.
      Set $a_L\leftarrow 0$ (leftmost), $a_R\leftarrow 0$ (rightmost).
\item \textbf{Increase tilts} until both endpoints dominate:
    \begin{enumerate}
    \item If $\widehat{\pi}_A(a_R) < \vartheta$: set $a_R \leftarrow a_R + s$, sample $n$ terminals at $a_R$, insert $a_R$ into $\mathcal{A}$.
    \item If $\widehat{\pi}_B(a_L) < \vartheta$: set $a_L \leftarrow a_L - s$, sample $n$ terminals at $a_L$, insert $a_L$ into $\mathcal{A}$.
    \end{enumerate}
\item \textbf{Ensure adjacent overlap:} Sort $\mathcal{A}$. While any neighboring pair $(a,a')$ violates
      $\widehat{\mathcal{O}}_{a\to a'}\!\ge\!\varepsilon_{\rm ov}$ \emph{or} $\widehat{\mathcal{O}}_{a'\to a}\!\ge\!\varepsilon_{\rm ov}$,
      insert the midpoint $\tilde a=(a+a')/2$, sample $n$ terminals at $\tilde a$, insert $\tilde a$ into $\mathcal{A}$, and update overlaps.
\item \textbf{Refine for accuracy:} Add more samples to all $a\in\mathcal{A}$ to reach ESS$_\mathrm{target}$.
\item \textbf{Unbias and estimate:} Solve weighted MBAR (Sec. \ref{subsec:weighted_MBAR}) over all tilts using bias potentials $b_a(x)=a(\xi(x)-\tfrac12)$ to obtain target ($a{=}0$) weights $W_n^{(0)}$, then compute
\begin{align}
\widehat{\Delta g}_{\mathrm{AB}}
\;&=\; -\log\frac{\sum_n W_n^{(0)}\,\xi(x_n)}{\sum_n W_n^{(0)}\,[1-\xi(x_n)]} \\
\widehat{\Delta G}_{\mathrm{AB}}&=k_BT\,\widehat{\Delta g}_{\mathrm{AB}}.
\end{align}
Uncertainties follow from the MBAR covariance or a cluster bootstrap over denoising-trajectory IDs.
\end{enumerate}

\paragraph{Illustration and results for protein folding with BioEmu}

\begin{figure*}
    \centering    \includegraphics[width=0.85\linewidth]{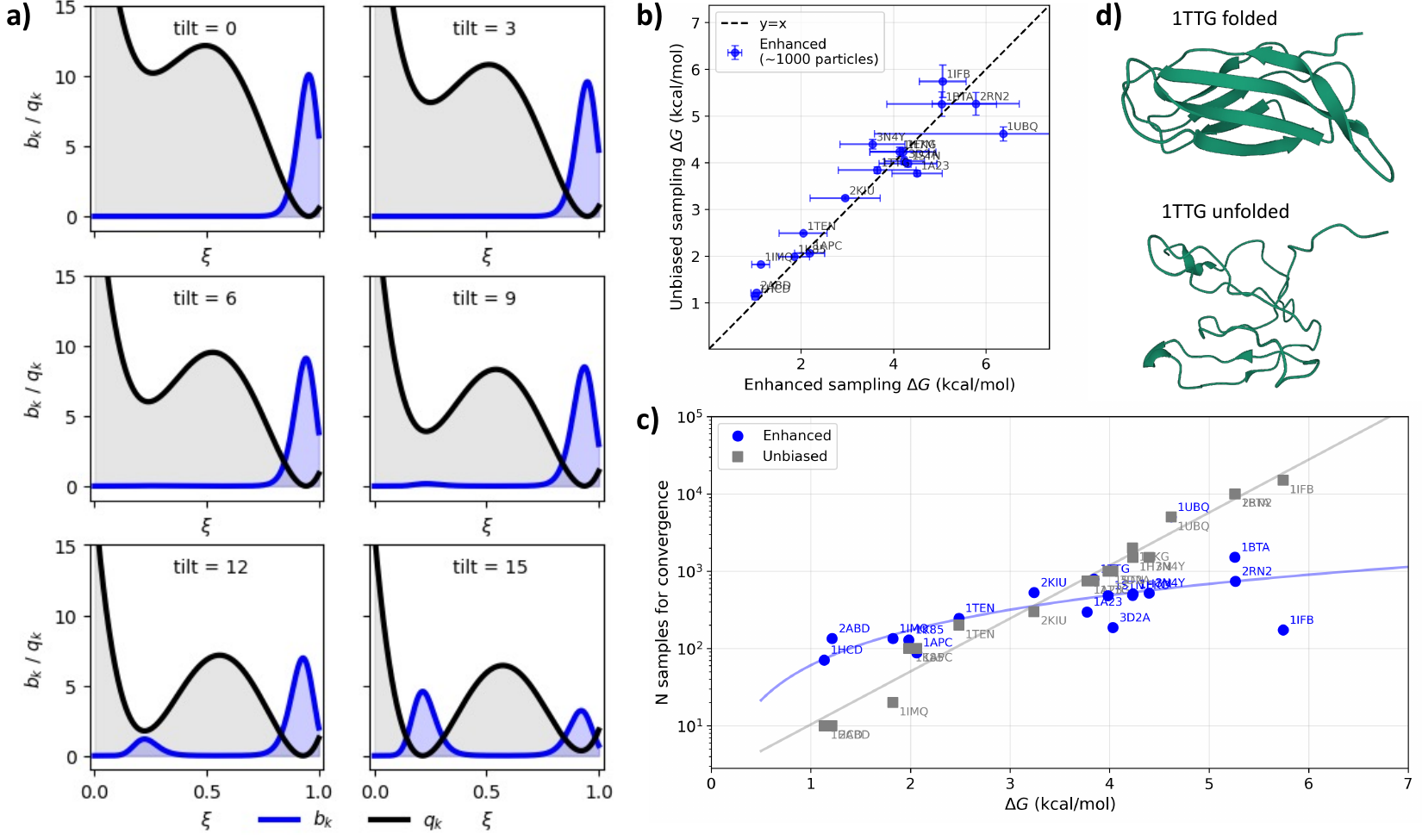}
    \caption{\textbf{$\Delta$G-Diff illustration and protein folding results with BioEmu}: a) Processes with large free energy differences (top left, black), e.g., protein folding, sample almost exclusively one state in equilibrium (top left, blue). $\Delta$G-Diff creates a series of tilted biased potentials until both states are dominantly sampled in at least one ensemble each (top left, bottom right), and these ensembles can be combined with MBAR to yield an estimate of $\Delta$G.
    b) Comparison of folding free energies $\Delta$G sampled by unsteered BioEmu and $\Delta$G-Diff with single optimal tilt. 
    c) Sampling efficiency as a function of $\Delta$G: unsteered sampling (grey) requires exponentially many samples in $\Delta$G, while the number of samples required by $\Delta$G-Diff shows much weaker scaling with $\Delta$G.
    d) Sample folded and unfolded state of a stable protein domain of fibronectin (PDB code 1TTG).}
    \label{fig:dGdiff_tilt_schematic}
\end{figure*}

Fig.~\ref{fig:dGdiff_tilt_schematic}a illustrates how $\Delta$G-Diff works for a two-state potential, which is representative of typical protein folding and protein-ligand binding scenarios. When the free energy difference between the two states $\Delta$G is large, almost all equilibrium density is in the more stable state, e.g., the folded or bound state
(Fig.~\ref{fig:dGdiff_tilt_schematic}a, top left).
$\Delta$G-Diff defines a series of tilt potentials interpolating between an ensemble in which one state is dominant (top left) and an ensemble in which the other state is dominant (bottom right). If a good guess of $\Delta$G is available, a single tilted ensemble which samples from both states is sufficient. The $\Delta$G-Diff ensembles can be generated with steering and combined with MBAR to the estimated free energy value. Note that in contrast to classical enhanced sampling methods that use MD, there is no need for having overlapping distributions in coordinate space. A single steered diffusion model ensemble that is able to sample both states (bottom right) is sufficient to obtain an unbiased estimate of $\Delta$G. Multiple ensembles are useful to find the optimal tilt, and combining their samples with MBAR improves statistics.

To connect these trends to practical compute, we implemented FKC reward tilting algorithm \cite{skreta2025feynman} in combination with the SDE DPM-Solver++ \cite{lu2025dpm}, a second order denoiser, based on the open source BioEmu \cite{bioemu2025} code. We used BioEmu model version 1.1 to generate samples of a set of relatively stable proteins selected from ProThermDB \cite{nikam2021prothermdb} database, with different predicted unbiased stabilities ranging from 1 to 6 kcal/mol. At room temperature, folding free energy of 5 kcal/mol corresponds to one unfolded sample in 4160 samples, which requires on the order of several GPU hours, with the exact timing depending on GPU and protein size. 
We have chosen estimated optimal tilts to illustrate the correctness and enhanced sampling efficiency for those proteins. 
For each protein, we generated 25,000 steered samples in total with small batches, using per-system optimized steering slopes along the collective variable defined as the Root Mean Squared Deviation (RMSD) to the native PDB structure. 
Unbiased statistics were recovered by direct reweighting (Eq.~\ref{eq:direct_reweighting}) since a single bias potential was used, and folding free energies were computed from the fraction of native contacts (FNC) using a two-state model. To quantify sample efficiency, since different batches are independently sampled, we repeatedly subsampled a given number of batches (and thus a fixed number of particles) without replacement, and evaluated the resulting $\Delta G$ estimates as a function of total sample count.

We excluded 8 systems where steering did not yield usable overlap, including 5 where the RMSD range is too large (beyond 2nm), and 3 where the unbiased reference estimates were not reliable due to unconverged sampling. After filtering, 18 proteins remain for quantitative comparison.
In Fig.~\ref{fig:dGdiff_tilt_schematic}b, we plot the enhanced sampling $\Delta G$ from reweighted steered samples at approximately 1,000 particles against the converged unbiased $\Delta G$, showing close agreement across the full stability range. 
We next measured the minimum sample count required for convergence, defining convergence as a self-reference mean absolute error (MAE) below 1 kcal/mol, as shown in Fig.~\ref{fig:dGdiff_tilt_schematic}c.
Ubiquitin (1UBQ) is a notable outlier, likely due to its non-two-state folding behavior.
For unbiased sampling, this required sample count grows exponentially with $\Delta G$, consistent with the fact that the probability of observing the thermodynamically rare state decreases exponentially in free energy difference. In contrast, steered sampling shows a much weaker scaling over the same range, demonstrating a substantial reduction in sample complexity for stable proteins.

The natural baselines for our folding free energy estimates would be converged MD-based references — either long unbiased trajectories or established enhanced sampling methods such as replica-exchange MD and metadynamics. However, obtaining such references for proteins in our stability range ($\Delta\text{G}_{\text{fold}}$ up to ~6 kcal/mol) requires millisecond-scale sampling accessible only on specialized hardware (e.g., Anton) or massive distributed simulation campaigns. We therefore reference each method against its own converged pooled estimate.
We also note concurrent work on diffusion-based steering \cite{NamEtAl_EnhancingDMSampling,RichmanDror_InferenceSteering,LamEtAl_Metadiffusion}, targeting simulation-free fine-tuning, hidden-state exploration, and experimental-data consistency, respectively, where the settings differ from ours.

Together, these experiments demonstrate the advantages of enhanced diffusion sampling. UmbrellaDiff and MetaDiff recombine biased ensembles via multi-state estimators into accurate free-energy surfaces, reducing sample requirements by one to several orders of magnitude. As a result, protein folding free energies scale sub-exponentially in $\Delta$ G, bringing previously intractable regimes within reach.

\section{Discussion}
Diffusion-model-based equilibrium samplers remove one of the two fundamental bottlenecks of molecular simulation: slow mixing due to time-correlated trajectories. However, iid sampling alone does not resolve the second bottleneck—the exponential sample complexity of estimating observables dominated by low-probability regions of the equilibrium distribution, such as in protein folding or protein-ligand binding. In this work, we show that the classical bias-and-reweighting paradigm of enhanced sampling can be transferred to diffusion samplers by steering pretrained models to generate biased ensembles and recovering unbiased thermodynamics with exact reweighting (direct reweighting, WHAM/MBAR).
In this sense, the present framework addresses both aspects of the sampling problem: diffusion models address slow mixing, and enhanced sampling addresses rarity, together enabling converged free-energy estimation on GPU timescales that were previously dominated by long MD trajectories.

A key consequence of replacing biased MD trajectories by independent (possibly weighted) samples is that several enhanced sampling procedures change character. In UmbrellaDiff, windows no longer need to be dynamically connected by slow transitions, and hidden barriers orthogonal to $\xi$ do not cause kinetic trapping within a window; overlap requirements are purely statistical and can be diagnosed and repaired using standard MBAR/WHAM tools (ESS, overlap matrices, uncertainty estimates). Likewise, in MetaDiff, history-dependent biasing becomes an \emph{equilibrium} sequence of thermodynamic states: each bias update defines a well-posed biased distribution that can be incorporated immediately into MBAR, allowing online diagnostics and early stopping without waiting for a fully “filled” surface. More broadly, once mixing is removed from the underlying sampling engine, bias variables need not correspond to slow dynamical modes, suggesting that reaction-coordinate learning and adaptive window placement may become simpler and more robust than in MD-based enhanced sampling.

At the same time, diffusion-based enhanced sampling inherits, and in some cases amplifies, several familiar limitations. 
First, the approach relies on having a sufficiently accurate pretrained equilibrium model for the molecular system and thermodynamic conditions of interest, and our results should be interpreted as enhanced sampling relative to the base diffusion-model distribution; any model mismatch between the pretrained model and physical thermodynamics propagates into reweighted estimates. 
Second, as in all importance-sampling schemes, weight degeneracy arises when biases are too aggressive or when the biased and target ensembles have insufficient overlap. This was the limiting factor for 5 of 26 candidate proteins in our BioEmu benchmark whose unfolded basins span RMSD ranges beyond 2 nm: steering pushes the score field into low-density regions of the pretrained model, producing extreme importance weights and degraded effective sample sizes. 
These exclusions motivate careful bias design, overlap diagnostics, and potentially additional intermediate states. 
Third, FKC steering requires SDE integrators; unbiased sampling can instead use ODE solvers. SDE schemes typically require more denoising steps than ODE schemes for equivalent fidelity \cite{holderrieth2025glass}, so the wall-clock costs reported here are specific to this setting and $\Delta G$ estimates carry a quantitative dependence on the integrator and step count.

Additionally, the present work focuses on equilibrium properties; extending these ideas to dynamical observables will require additional structure, for example through path reweighting or consistent generative dynamical models.

Looking forward, several directions appear particularly promising. Methodologically, the steering framework can be combined with adaptive schemes that learn reaction coordinates or optimal bias potentials on the fly, and with automated window/tilt construction guided by MBAR diagnostics. Architecturally, energy-based or hybrid diffusion models that allow direct evaluation of $u(x)$ could enable tighter estimator control and new forms of biasing.

Beyond equilibrium ensemble generators, an active line of research learns accelerated \emph{dynamics} or \emph{transport operators} from MD data, including Timewarp \cite{KleinEtAl_Timewarp}, implicit transfer operator learning \cite{SchreinerWintherOlsson_ITO}, and trajectory generators such as MDGen \cite{JingStaerkJaakkolaBerger_MDGen}.

Because these models are trained to reproduce finite-lag transitions rather than the equilibrium distribution, their stationary distribution need not coincide with the desired $p(x)$ unless additional constraints or corrections are imposed. 

One promising route is to wrap learned transport models into Metropolis--Hastings or related accept/reject schemes that target a known equilibrium distribution (as done in Timewarp), in which case bias potentials can be incorporated by changing the target and standard reweighting estimators apply. 
Another route is to extend biasing and reweighting to \emph{path ensembles}, where one biases trajectory functionals (e.g., for transition-path sampling) and reweights by likelihood ratios of paths; MDGen already highlights transition-path sampling as a target application. 
Recent work on incorporating equilibrium priors into transfer-operator learning (e.g., BoPITO \cite{VigueraDiezEtAl_BoPITO}) further suggests opportunities to unify learned transport with thermodynamically consistent sampling.

Finally, while we have emphasized biomolecular folding and conformational change, the formulation is general and applies to any system where iid equilibrium samplers exist but rare-event statistics are the remaining bottleneck—including materials, soft matter, and condensed-phase chemistry. We anticipate that integrating enhanced sampling principles into diffusion samplers will become a central building block for next-generation molecular simulation workflows, enabling routine computation of free energies and rare-state observables at scales previously accessible only with specialized hardware or massive distributed MD.

\section*{Code availability}

The BioEmu inference code is in \url{https://github.com/microsoft/bioemu}, with the FKC steering feature and enhanced sampling examples on Gaussian mixture models (GMMs) and real proteins.

\section*{Acknowledgments}
This paper is dedicated to Gerhard Hummer on the occasion of his 60th birthday. Dear Gerhard, happy birthday and thank you for your many contributions and for being a constant source of inspiration for the field!

We acknowledge Carles Domingo-Enrich and Akshay Krishnamurthy for helpful discussions. GitHub Copilot has been used in part of the implementation and analysis of this work. This work has been funded by Microsoft Research.


\bibliography{all,own}


\appendix

\section{BioEmu Folding Free Energies Details}

We selected 26 proteins from the ProThermDB database, ranging from 76 to 372 residues. The steering potential is a clamped linear function of the collective variable, $V = s \cdot \mathrm{clamp}(\mathrm{CV}, c_\mathrm{min}, c_\mathrm{max})$, where $s$ is a per-system estimated optimal slope, and $c_\mathrm{min/max}$ are the clipping bounds to prevent the potential going unbounded. Within each run, the steered samples are generated in small independent batches with batch size dependent of the protein size, allowing us to pool and subsample batches across runs to study convergence at varying sample sizes.
Specifically, we randomly draw (without replacement) subsets of batches from the combined pool and compute $\Delta G$ for each draw, repeating 200 times per sample size to obtain statistics. We note the subsampling is performed on the batch level because different batches are independent, but not within each batch because samples in the same batch have already been resampled through the steering algorithm.
Folding free energies are estimated from the fraction of native contacts (FNC) using a two-state model with inverse Boltzmann reweighting to correct for the steering bias.

\begin{figure*}[htbp]
    \centering
    \includegraphics[width=0.9\linewidth]{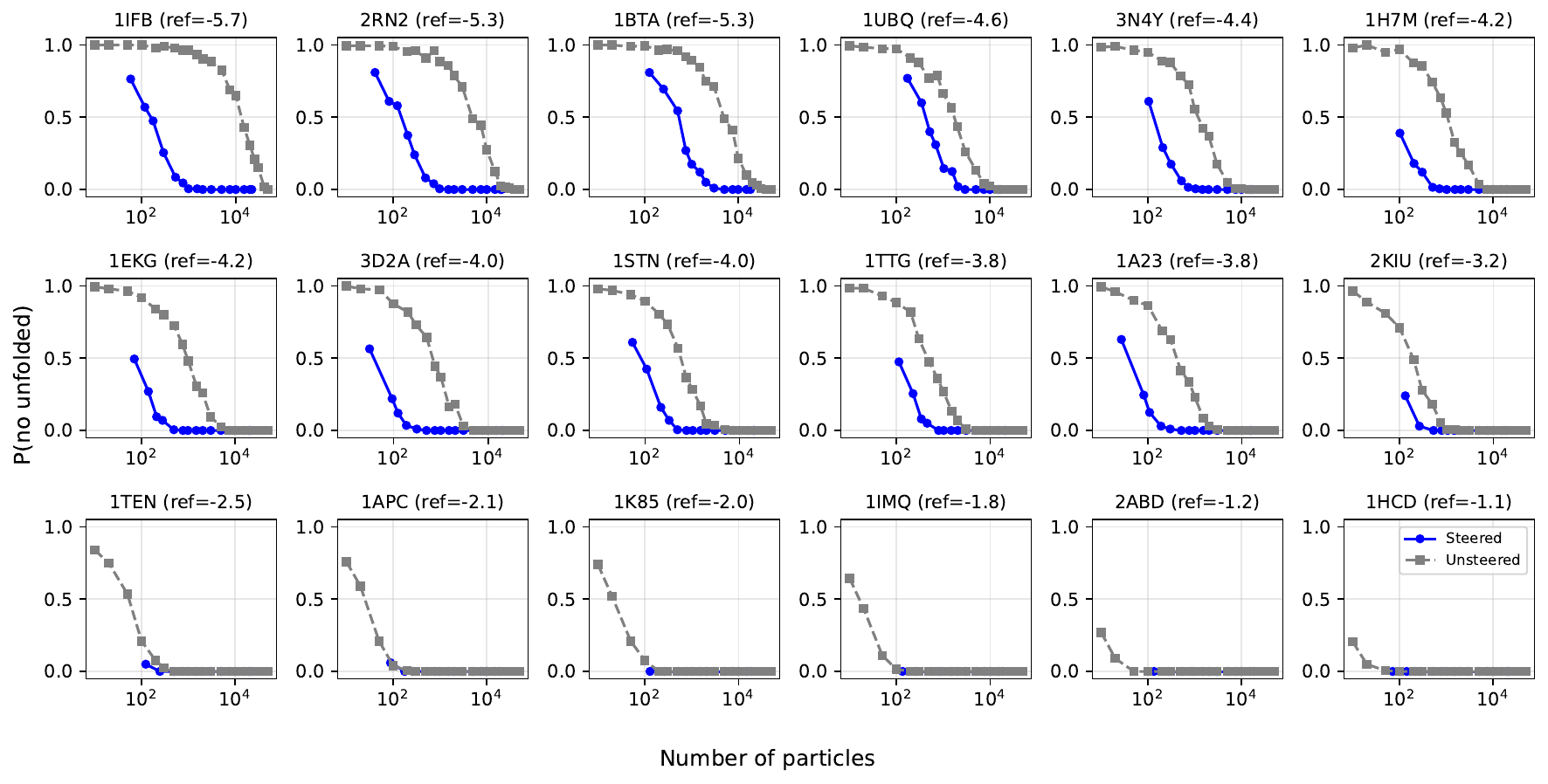}
    \caption{Per-system catastrophic failure rate as a function of sample size for 18 ProThermDB proteins, sorted by reference $\Delta G$. Catastrophic failure is defined as the probability that a random draw of $n$ particles contains no unfolded sample (FNC $< 0.5$). Blue: steered sampling; grey: unbiased sampling. Steered sampling eliminates catastrophic failures at small sample sizes across all systems, while unbiased sampling requires orders of magnitude more particles for very stable proteins.}
    \label{app-fig:catastrophic_failure}
\end{figure*}

\begin{figure*}[htbp]
    \centering
    \includegraphics[width=0.9\linewidth]{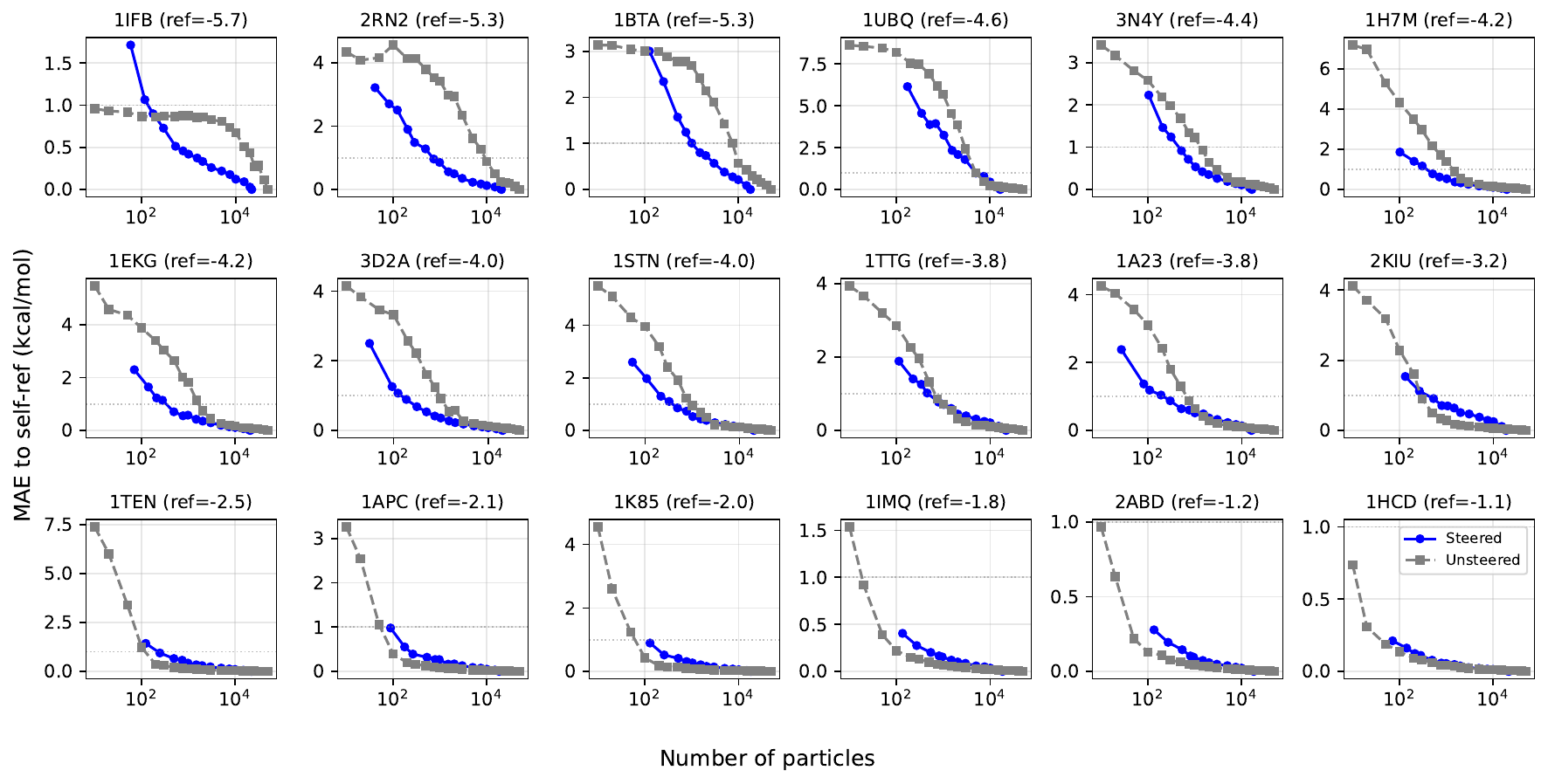}
    \caption{Per-system mean absolute error (MAE) of $\Delta G$ as a function of sample size, with each method referenced to its own converged pooled estimate. Proteins are sorted by reference $\Delta G$ (most stable at top-left). Steered sampling (blue) converges substantially faster than unbiased sampling (grey), with the advantage most pronounced for more stable systems.}
    \label{app-fig:mae_selfref}
\end{figure*}

We exclude 8 of the 26 systems from the main analysis: 5 proteins (1QLP, 1CEC, 2LZM, 1RHD, 1RIL) whose RMSD ranges beyond 2nm, leading to difficult steering and extreme reweighting factors, and 3 proteins (1HK0, 1IOB, 1Y9O) for which the unbiased reference $\Delta G$ is unreliable due to unconverged sampling of the unfolded state. The remaining 18 systems (1A23, 1APC, 1BTA, 1EKG, 1H7M, 1HCD, 1IFB, 1IMQ, 1K85, 1STN, 1TEN, 1TTG, 1UBQ, 2ABD, 2KIU, 2RN2, 3D2A, 3N4Y) span predicted unbiased $|\Delta G_{\mathrm{fold}}|$ values from 1.1 to 5.7 kcal/mol. 
Fig.~\ref{app-fig:catastrophic_failure} shows the catastrophic failure rate—defined as the probability that a random draw of $n$ particles contains no unfolded sample (FNC $< 0.5$), as a function of sample size for each system. For steered sampling, this rate drops to zero at modest sample sizes across nearly all systems, whereas unbiased sampling requires substantially more particles, particularly for thermodynamically stable proteins. Fig.~\ref{app-fig:mae_selfref} shows the corresponding mean absolute error (MAE) of $\Delta G$ relative to each method's own converged estimate, as a function of sample size. Steered sampling achieves sub-kcal/mol accuracy at $n \approx 100$-$1{,}000$ depending on stability. While unbiased sampling converges fast on systems with relatively small $\Delta G$, the required number of samples grow exponentially with $\Delta G$. The steered sampling outperforms the unsteered sampling at $\Delta G \approx 3.2$ kcal/mol, and the advantage becomes more significant with larger $\Delta G$.

\end{document}